\documentclass[lettersize,journal]{IEEEtran}
\usepackage[utf8]{inputenc}
\usepackage{amsmath,amsfonts}
\usepackage{algorithmic}
\usepackage{algorithm}
\usepackage{array}
\usepackage[caption=false,font=normalsize,labelfont=sf,textfont=sf]{subfig}
\usepackage{textcomp}
\usepackage{stfloats}
\usepackage{url}
\usepackage{verbatim}
\usepackage{graphicx}
\usepackage{cite}
\usepackage{tabularx}
\usepackage{amsmath}
\usepackage{booktabs} 
\usepackage{tabularx}
\usepackage{array}
\usepackage{pgfplots} 
\pgfplotsset{compat=1.18}
\usepackage[hidelinks, colorlinks]{hyperref}
\usepackage{xcolor}
\usepackage[table]{xcolor}
\usepackage{multirow}   
\usepackage{colortbl}

\begin{document}

\title{
Obshazard-bench: Benchmarking Multimodal Foundation Models for Real-Time Disaster Intelligence from Raw Earth Observation Streams
}
        
\author{
Fengxiang Wang$^{1}$,
Qiuyang Yu$^{1}$,
Yueying Li$^{1}$,
Mingshuo Chen$^{1}$,
Chengchi Fei$^{2}$,
Kaiyi Xu$^{2}$,
Lixin Gu$^{2}$,
Wangxu Wei$^{2}$,
Junchao Gong$^{2}$,
Lipeng Ma$^{2}$,
Jiong Wang$^{2}$,
Fenghua Ling$^{2}$,
Wenlong Zhang$^{2}$,
Xue Yang$^{3}$,
Wenjing Yang$^{1}$,
Ben Fei$^{2,*}$,
Long Lan$^{1,*}$\IEEEmembership{Member,~IEEE}

\thanks{
$^{1}$ National University of Defense Technology, China.

$^{2}$ Shanghai Artificial Intelligence Laboratory, China.

$^{3}$ Shanghai Jiao Tong University, China.

* Corresponding authors: Ben Fei and Long Lan.
  }

}

\markboth{Journal of \LaTeX\ Class Files,~Vol.~14, No.~8, August~2021}%
{Shell \MakeLowercase{\textit{et al.}}: A Sample Article Using IEEEtran.cls for IEEE Journals}


\maketitle

\begin{abstract}
Multimodal Large Language Models (MLLMs) are increasingly used to interpret Earth observation data, yet their capability to support real-world disaster emergency response remains insufficiently evaluated.
Existing remote sensing benchmarks largely rely on static, post-hoc, and expert-processed products, such as gridded reanalysis data, which are difficult to align with operational disaster scenarios where hazards evolve rapidly and decisions must be made under strict time constraints.
To bridge this gap, we introduce Obshazard-bench, a real-time, observation-driven benchmark for evaluating disaster intelligence in MLLMs.
Unlike image-centric or post-event benchmarks, Obshazard-bench directly integrates raw, high-frequency satellite sounding streams from diverse satellite sensors with concurrent ground-station observations, historical disaster records, and socio-economic indicators, bypassing delayed expert-processing and physical-inversion pipelines.
The benchmark covers 8 major disaster categories and 28 sub-categories across more than 60 countries, incorporating over 120 historically documented extreme-event cases and thousands of lifecycle-oriented VQA samples.
Moreover, Obshazard-bench further defines a three-stage evaluation taxonomy aligned with the operational disaster workflow: Predictive Crisis Anticipation for pre-disaster risk detection and early forecasting, Active Evolution Reasoning for in-situ disaster tracking and termination prediction, and Multi-faceted Impact Quantification for post-disaster magnitude deduction, humanitarian burden estimation, and socio-economic impact assessment.
Experiments on representative general-purpose and Earth-focused foundation models reveal substantial limitations in transforming raw multi-channel physical observations into temporally grounded and decision-relevant disaster reasoning. The dataset and evaluation code are available at: \href{https://github.com/YYQ898/Obshazard-bench}{\color{magenta}Obshazard-bench}
\end{abstract}

\begin{figure*}[t]
    \centering

    \includegraphics[width=1.0\textwidth]{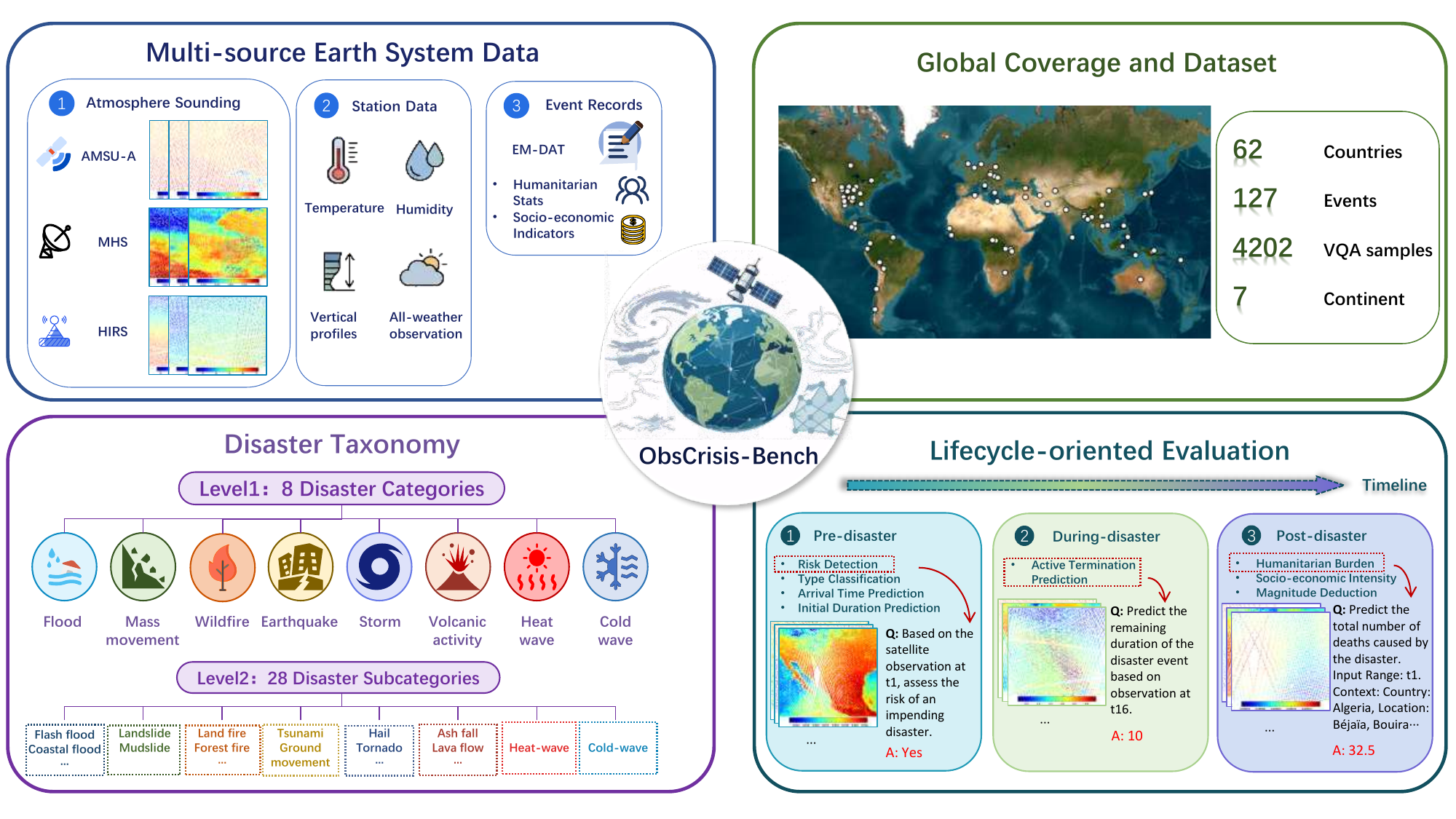}
    \caption{\textbf{System architecture of the proposed Disaster Intelligence Benchmark} (conceptualized in Section~\ref{sec:motivation} and structured in Section~\ref{sec:benchmark}). The framework integrates: (Top-Left) Multi-source Earth system data coupling raw atmospheric sounding streams with ground measurements; (Top-Right) Global scale distribution covering over 60 countries; (Bottom-Left) Two-level hierarchical hazard taxonomy; and (Bottom-Right) Three-stage disaster lifecycle-oriented evaluation supported by representative visual question-answering examples.}
    \label{fig:main-logo} 
\end{figure*}

\section{Introduction}
\label{sec:motivation}

\begin{figure*}[!t]
	\centering            
        \includegraphics[width=\linewidth]{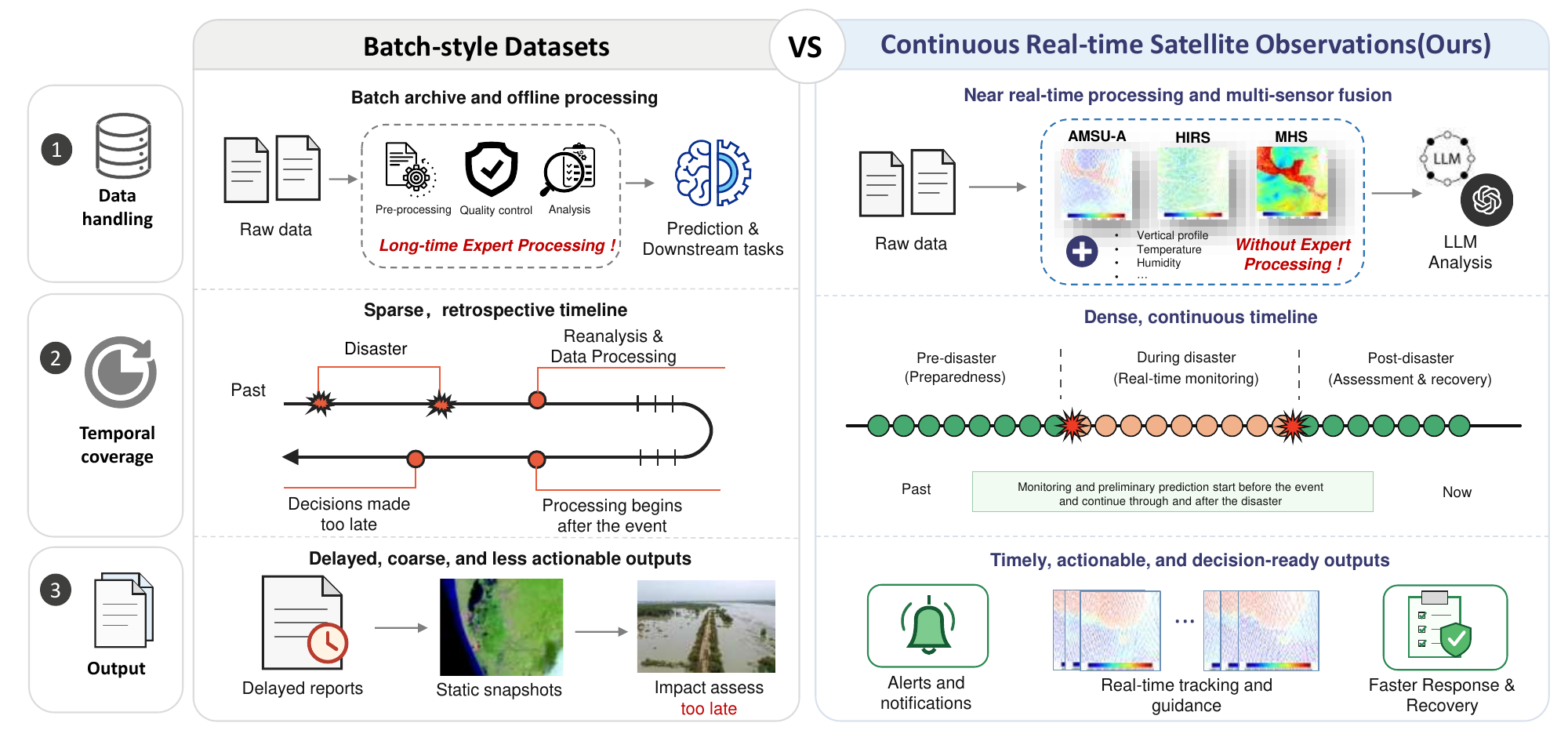}
    \caption{\textbf{Conceptual comparison between traditional batch-style evaluation and our proposed real-time streaming paradigm.} The comparison is organized across three critical levels: (Row 1) \textbf{Data Handling,} contrasting delayed offline archives with direct, raw satellite feeds (AMSU-A, HIRS, MHS) bypassing expert processing; (Row 2) \textbf{Temporal Coverage,} highlighting sparse retrospective analysis versus dense, continuous lifecycle monitoring; and (Row 3) \textbf{Operational Decision Utility,} showing how our paradigm enables proactive \textbf{alerts and notifications} pre-event, continuous \textbf{real-time tracking and guidance} in-situ, and \textbf{faster response and recovery} strategies post-event.}
    \label{fig:data_type}   
    \vspace{-3mm}
\end{figure*}

Recent breakthroughs in Multimodal Large Language Models (MLLMs) and Earth Foundation Models (EFMs) have advanced spatial-temporal reasoning for Earth observation (EO) data, spanning multi-modal geospatial understanding \cite{kuckreja2024geochat, zhang2024earthgptuniversalmultimodallarge, guo2024skysense, hong2024spectralgpt2, bell2026earthaiunlockinggeospatial} and global atmospheric forecasting \cite{bi2023accurate, lam2023learning, nguyen2023climax, chen2023fengwu, jakubik2023foundation}. Evaluating these systems requires rigorous benchmarking. While existing benchmarks like XLRS-Bench \cite{wang2025xlrs}, OmniEarth-Bench \cite{chen2026omniearth}, VRSBench \cite{vrsbench2024}, and RSRSD-5M \cite{shen2026rsrsd} assess various remote sensing (RS) vision tasks, evaluating MLLMs in real-world disaster contexts remains difficult due to rapid event evolution and complex physical variables. Specially, existing benchmarks are limited in three main areas:

\paragraph{\textbf{Operational Processing Latency}} 
Disaster emergency response requires tracking atmospheric and surface hazards that evolve rapidly in real time. For timely warnings, MLLMs must process continuous, live, and high-frequency observation streams \cite{andrychowicz2023deep}. However, current benchmarks mostly use static, expert-processed datasets like orthorectified imagery or gridded reanalysis products. These products require time-consuming steps like expert cleaning and physical inversion, which introduce delays of days or weeks (Fig.~\ref{fig:data_type}). This latency makes them impractical for active emergency response.

\paragraph{\textbf{Geographical and Hazard Scope}} 
Natural disasters vary widely across different climates and regions. To test model generalization, a benchmark needs global coverage with diverse hazard types. Many existing disaster datasets focus on specific areas or limited hazard types, often concentrating on post-event structural damage.

\paragraph{\textbf{Disaster Lifecycle-Oriented Evaluation}} 
Emergency decisions are made continuously as a disaster develops. However, evaluating models only on post-event imagery isolates the event from its temporal context. Practical disaster management requires MLLM that can perform reasoning throughout the event lifecycle. Existing datasets do not support this temporal dependency. 

To address these challenges, we introduce a real-time, observation-driven disaster intelligence benchmark that integrates continuous satellite observations, ground station data, authentic extreme-event records, and socio-economic indicators. Instead of using pre-processed visual maps, our framework replicates the professional meteorologist's workflow by directly providing raw, multi-channel physical sounding profiles from core weather satellite sensors: the Advanced Micowave Sounding Unit-A (AMSU-A, 15 channels), the High-Resolution Infrared Radiation Sounder (HIRS, 20 channels), and the Microwave Humidity Sounder (MHS, 5 channels) \cite{camps-valls_artificial_2025, reichstein2025early}. In operational centers, meteorologists analyze these exact vertical sounding profiles to detect atmospheric instability and moisture transport. Spanning 8 major disaster categories and 28 sub-categories across more than 60 countries (encompassing over 120 historical extreme-event cases), our benchmark evaluates whether foundation models can directly ingest raw sensor streams and transform them into structured, decision-relevant reasoning across three progressive lifecycle phases: pre-event anomaly detection, during-event intensity tracking, and post-event consequence reasoning. Ultimately, this setup evaluates whether MLLMs can bypass delayed processing pipelines to support practical decision-making across the disaster lifecycle---from issuing early alerts to guiding real-time tracking and accelerating post-event recovery (Fig.~\ref{fig:data_type}).

\begin{table}[t!]
\caption{Quantitative Comparison with Existing Disaster-Specific Benchmarks}
\label{tab:benchmark_comparison}
\centering

\resizebox{0.48\textwidth}{!}{\begin{tabular}{lccccc}
\toprule
\textbf{Dataset} & 
\shortstack{\textbf{Disaster}\\\textbf{Types}} & 
\shortstack{\textbf{Real}\\\textbf{Events}} & 
\shortstack{\textbf{VQA}\\\textbf{Samples}} & 
\shortstack{\textbf{Covered}\\\textbf{Countries}} & 
\shortstack{\textbf{Raw}\\\textbf{Stream}} \\
\midrule
CrisisMMD \cite{alam2018crisismmd} & 7 & 7 & -- & 5 & $\times$ \\
xBD \cite{gupta2019xbd} & 6 & 19 & -- & 6 & $\times$ \\
Sen12Floods \cite{bonafilia2020sen12floods} & 1 & 11 & -- & 11 & $\times$ \\
FloodNet \cite{rahnemoonfar2021floodnet} & 1 & 1 & 4,500 & 1 & $\times$ \\
RescueNet \cite{gupta2020rescuenetjointbuildingsegmentation} & 1 & 1 & -- & 1 & $\times$ \\
CRASAR-U-D. \cite{manzini2024crasar} & 2 & 10 & -- & 1 & $\times$ \\
DisasterM3 \cite{wang2025disasterm3} & 10 & 36 & 26,988 & 5 & $\times$ \\
ZeShot-VQA \cite{karimi2025zeshot} & 4 & 19 & 15,000 & 6 & $\times$ \\
DisasterVQA \cite{almohannadi2026disastervqa} & 13 & 15 & 4,405 & 1 & $\times$ \\
DORA \cite{wang2026dora} & 10 & 45 & 515 & 5 & $\times$ \\
BRIGHT \cite{chen2025bright} & 7 & 14 & -- & 14 & $\times$ \\
EBD \cite{wang2025ebd} & 12 & 12 & -- & 8 & $\times$ \\
MONITRS \cite{monitrs2025} & 8 & 9,996 & -- & 1 & $\times$ \\
\midrule
\textbf{Obshazard-bench} & \textbf{28} & \textbf{127} & \textbf{4,202} & \textbf{62} & \textbf{$\checkmark$} \\
\bottomrule
\end{tabular}}
\vspace{-2mm} 
\end{table}

The main contributions of this paper are summarized as follows:
\begin{itemize}
    \item \textbf{Direct Sensor-Stream Evaluation Paradigm:} We introduce a disaster benchmark built on raw, live-streaming satellite and ground data, bypassing the delayed processing pipelines of traditional reanalysis datasets.
    \item \textbf{Global-scale Multi-hazard Testbed:} The benchmark provides global coverage with 8 major disaster categories and 28 sub-categories across more than 60 countries, incorporating over 120 historical extreme-event cases to test model generalization.
    \item \textbf{Lifecycle-aligned Evaluation Taxonomy:} We organize tasks into a three-stage temporal framework spanning pre-event predictive anticipation, during-event active evolution tracking, and post-event impact quantification, enabling continuous evaluation.
    \item \textbf{Systematic Functional Gap Assessment:} We evaluate several open- and closed-source foundation models on our benchmark tasks. Our analysis systematically exposes their functional deficiencies and performance bottlenecks in translating raw, multi-channel physical signals into temporal and logical reasoning, helping identify key limitations in driving practical disaster mitigation applications.
\end{itemize}

\section{Related Work}
\label{sec:related}

\subsection{Foundational Remote Sensing Datasets and Benchmarks}

Remote sensing (RS) datasets have evolved from early archival time-series analysis~\cite{zhu2017landsat, mayr2019validation} to large-scale multimodal benchmarks. Recent works such as XLRS-Bench~\cite{wang2025xlrs}, VRSBench~\cite{vrsbench2024}, and RSRSD-5M~\cite{shen2026rsrsd} have advanced ultra-high-resolution (UHR) remote sensing perception. However, most of these datasets are still built around static optical imagery. This limits their use in disaster scenarios, where cloud interference is common and where atmospheric structure cannot be captured from surface appearance alone~\cite{zhang2026benchmark, fu2024remote}.

Several benchmarks have started to include temporal or multi-modal observations. DynamicEarthNet~\cite{toker2022dynamic} provides daily multi-spectral sequences, and FoMo-Bench~\cite{rolf2025fomo} introduces multi-modal forest monitoring. Other specialized datasets, such as TreeFinder~\cite{wang2025treefinder} and SegMunich in SpectralGPT~\cite{hong2024spectralgpt2}, further enrich remote sensing data coverage. Yet these datasets mainly describe surface land-cover or vegetation changes. They do not focus on fast-changing physical processes in the atmosphere, which are critical for many extreme events.

Recent Earth foundation models and benchmarks also point to a broader trend toward cross-sphere and heterogeneous Earth-system reasoning. Examples include AnySat~\cite{astruc2025anysat}, TerraMind~\cite{jakubik2025terramind}, OmniEarth-Bench~\cite{chen2026omniearth}, OmniGAIA~\cite{li2026omnigaia}, Earth-Agent~\cite{feng2025earth}, and TerraBench~\cite{nguyen2026terrabench}. These works expand Earth AI evaluation beyond isolated visual perception, covering broader data sources, cross-domain reasoning, and in some cases tool-augmented agent workflows. However, high-frequency physical observations, such as vertical temperature and humidity sounding signals, are still underrepresented in benchmark design. Existing benchmarks also rarely organize such observations around disaster response stages. Obshazard-bench complements these efforts by focusing on raw sounding streams and lifecycle-oriented disaster reasoning.

Obshazard-bench addresses this gap by using high-frequency sounding streams from AMSU-A, HIRS, and MHS. These instruments provide multi-channel physical observations of the atmosphere, including AMSU-A with 15 channels and HIRS with 20 channels. Compared with optical-centric benchmarks, Obshazard-bench offers a complementary physical view of disasters. It allows models to reason over atmospheric signals that are not visible in traditional surface imagery.

\subsection{Benchmarks for Extreme Disaster Response}

Remote sensing benchmarks for disaster response have moved from basic perception toward more complex reasoning. Early datasets such as xBD~\cite{gupta2019xbd}, RescueNet~\cite{rahnemoonfar2023rescuenet}, and CRASAR-U-DRoIDs~\cite{manzini2024crasar} mainly support building damage assessment from bi-temporal satellite images or unmanned aerial vehicle (UAV) imagery. These datasets are important for post-disaster mapping, but they are less suited to early warning or event evolution analysis.

Later benchmarks extend the task scope and hazard coverage. DisasterM3~\cite{wang2025disasterm3} and RSCC~\cite{chen2025rscc} introduce multi-modal reasoning and change captioning. ZeShot-VQA~\cite{karimi2025zeshot} and DisasterVQA~\cite{almohannadi2026disastervqa} explore disaster-related visual question answering. Other works, including Anomaly-CD~\cite{li2024anomaly}, Shield~\cite{beijing2026shield}, BRIGHT~\cite{chen2025bright}, EBD~\cite{wang2025ebd}, and MONITRS~\cite{monitrs2025}, further improve multi-hazard coverage and dataset scale.

Despite these advances, most disaster benchmarks still rely on static or bi-temporal imagery. Their tasks are often centered on damage classification, change detection, or basic VQA. They rarely use raw high-frequency physical streams, and they do not fully reflect the operational needs of disaster response. In practice, disaster management requires more than post-event perception. Models must support early warning, track event evolution, and estimate humanitarian or socio-economic impacts.

Meteorological foundation models provide useful evidence that high-frequency physical data can support such reasoning. Models such as Aurora~\cite{bodnar2025aurora}, Aardvark Weather~\cite{allen2025aardvark}, and Prithvi WxC~\cite{mukkavilli2023prithvi} show strong potential for weather forecasting and early warning. However, these works mainly focus on model development and prediction tasks. They are not designed as benchmarks for evaluating multimodal disaster intelligence across multiple hazards and lifecycle stages.

This leaves an operational gap between experimental Earth AI systems and professional disaster management~\cite{reichstein2025early, bell2026earthaiunlockinggeospatial}. Recent reviews also note that current systems still struggle to assimilate high-frequency physical data for response-oriented reasoning~\cite{kim_ai_2025, preisser2025remote, camps-valls_artificial_2025}. Obshazard-bench is designed to address this gap. It covers 8 major disaster categories and 28 sub-categories across 62 countries. It aligns daily physical observation sequences with 4,202 lifecycle-oriented VQA samples, providing a realistic testbed for evaluating disaster-oriented MLLMs.

\section{Obshazard-bench}
\label{sec:benchmark}

\subsection{Data Sources and Processing}
\label{subsec:data}

Obshazard-bench is built upon a multi-source architecture that connects verified disaster records with high-frequency geophysical observations. We use the EM-DAT international disaster database~\footnote{https://www.emdat.be/} as the event metadata source, from which historical disaster cases are selected and organized. To focus on rapidly evolving extreme events, we retain disasters with an active duration no longer than 9 days. The resulting benchmark contains 127 verified historical extreme-event cases across 62 countries, covering 8 major disaster categories and 28 disaster sub-categories.

Departing from conventional optical-centric or post-event disaster benchmarks, Obshazard-bench uses raw satellite sounding streams as the core observational input. Specifically, we incorporate a complementary tri-sensor constellation consisting of AMSU-A with 15 channels, HIRS with 20 channels, and MHS with 5 channels. These instruments provide multi-channel atmospheric sounding observations related to temperature, humidity, and other physical conditions, enabling models to reason from raw geophysical signals rather than relying only on surface-level visual appearance. In addition to satellite streams, the benchmark integrates disaster metadata, ground-station observations, historical disaster records, and socio-economic indicators when constructing lifecycle-oriented disaster reasoning tasks.

The processing procedure is designed to preserve the raw-stream nature of the observations while making heterogeneous sources usable for multimodal evaluation. For each selected disaster event, satellite observations are temporally aligned with the event window and spatially associated with the affected region. The benchmark does not replace raw observations with delayed reanalysis products, expert-interpreted maps, or physically inverted variables. Instead, it organizes raw multi-channel sensor observations into event-centric samples, allowing Obshazard-bench to evaluate whether MLLMs can transform raw physical observations into temporally grounded and decision-relevant disaster reasoning.

\begin{table}[t]
\centering
\scriptsize
\setlength{\tabcolsep}{3pt}
\renewcommand{\arraystretch}{1.05}
\caption{Disaster taxonomy and sample distribution (single column).}
\label{tab:disaster}
\begin{tabularx}{\columnwidth}{>{\raggedright\arraybackslash}X >{\raggedright\arraybackslash}X r r r}
\toprule
\rowcolor{gray!15}
\textbf{L1 Category} & \textbf{L2 Sub-categories} & \textbf{Samples} & \textbf{Ratio} & \textbf{Events} \\
\midrule
\multirow{4}{*}{Flood} & Flash flood & 95 & 2.3\% & 5 \\
                        & Coastal flood & 68 & 1.6\% & 2 \\
                        & General flood & 195 & 4.6\% & 5 \\
                        & Riverine flood & 165 & 3.9\% & 5 \\
\midrule
\multirow{2}{*}{Mass movement (wet)} & Landslide & 66 & 1.6\% & 2 \\
                                      & Mudslide & 123 & 2.9\% & 4 \\
\midrule
\multirow{3}{*}{Wildfire} & Land fire & 170 & 4.0\% & 5 \\
                          & Forest fire & 159 & 3.8\% & 5 \\
                          & General wildfire & 137 & 3.3\% & 4 \\
\midrule
\multirow{2}{*}{Earthquake} & Tsunami & 200 & 4.8\% & 5 \\
                            & Ground movement & 217 & 5.2\% & 5 \\
\midrule
\multirow{11}{*}{Storm} & Hail & 174 & 4.1\% & 5 \\
                        & Tornado & 195 & 4.6\% & 5 \\
                        & Blizzard/Winter storm & 145 & 3.5\% & 5 \\
                        & Derecho & 128 & 3.0\% & 5 \\
                        & Extra-tropical storm & 150 & 3.6\% & 5 \\
                        & General storm & 184 & 4.4\% & 5 \\
                        & Lightning/Thunderstorm & 160 & 3.8\% & 5 \\
                        & Sand/Dust storm & 130 & 3.1\% & 5 \\
                        & Severe weather & 190 & 4.5\% & 5 \\
                        & Storm surge & 191 & 4.5\% & 5 \\
                        & Tropical cyclone & 205 & 4.9\% & 5 \\
\midrule
\multirow{4}{*}{Volcanic activity} & Ash fall & 135 & 3.2\% & 5 \\
                                   & Lava flow & 123 & 2.9\% & 4 \\
                                   & General activity & 163 & 3.9\% & 5 \\
                                   & Pyroclastic flow & 29 & 0.7\% & 1 \\
\midrule
\multirow{1}{*}{Heat-wave} & Heat-wave & 165 & 3.9\% & 5 \\
\midrule
\multirow{1}{*}{Cold-wave} & Cold-wave & 140 & 3.3\% & 5 \\
\midrule
\textbf{Total} & \textbf{28 sub-categories} & \textbf{4,202} & \textbf{100.0\%} & \textbf{127} \\
\bottomrule
\end{tabularx}
\end{table}

\subsection{Benchmark Construction Pipeline}
\label{subsec:pipeline}

\begin{figure*}[!t]
    \centering
    \includegraphics[width=\linewidth]{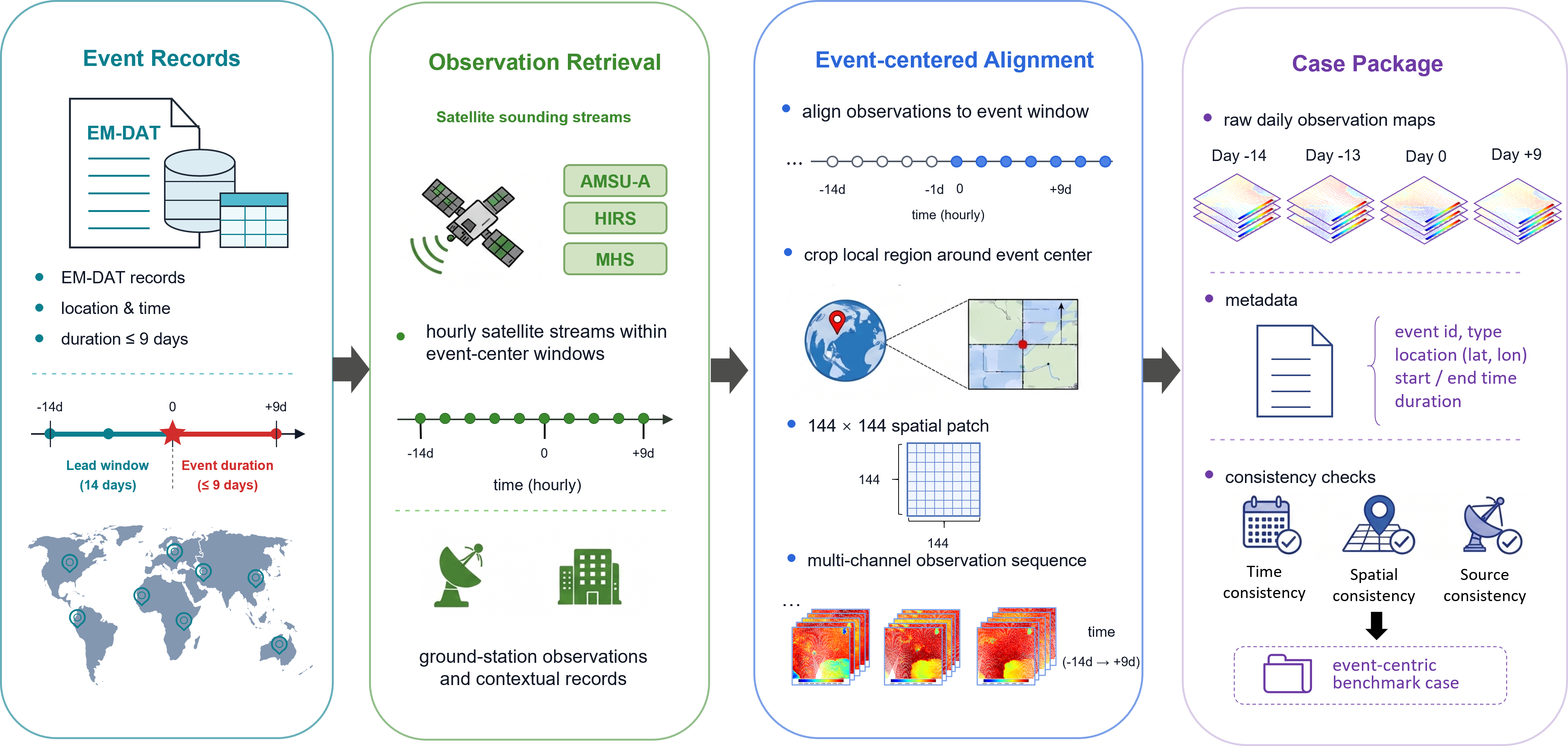}
    \caption{\textbf{Benchmark construction pipeline of Obshazard-bench.}
    The benchmark is constructed through four stages: (1) disaster event screening from EM-DAT records with spatial and temporal metadata, (2) retrieval of multi-source observations including AMSU-A, HIRS, MHS, and auxiliary contextual records, (3) event-centered spatial-temporal alignment that associates observations with disaster locations and time windows, and (4) construction of standardized benchmark cases containing observation sequences, metadata, and quality-control information. This pipeline preserves the raw-stream characteristics of satellite sounding observations while organizing heterogeneous sources into event-centric benchmark instances.}
    \label{fig:pipeline}
    \vspace{-3mm}
\end{figure*}

Obshazard-bench is constructed through a multi-stage pipeline that transforms heterogeneous disaster records and raw satellite observations into reasoning-oriented benchmark samples. The pipeline begins with disaster event metadata. For each event, we extract the disaster identifier, start date, end date, longitude, latitude, and disaster type. Events with durations longer than 9 days are filtered out, ensuring that the benchmark focuses on rapidly evolving disaster processes. To support pre-disaster reasoning, the observation window is extended by shifting the start date 14 days earlier than the recorded event onset.

For each retained event, raw satellite observations are collected within the corresponding temporal window. The current benchmark uses AMSU-A, HIRS, and MHS observations. For each day in the event-centered time span, the pipeline reads observations within a 24-hour window, typically from 00:00 to 23:00 at hourly intervals. These observations are retrieved from the storage system according to the corresponding sensor type, time stamp, and event location.

After temporal retrieval, the observations are spatially cropped around the event center. Given the event longitude and latitude, a local region is extracted using a padding size of 72 pixels, resulting in a $144 \times 144$ spatial crop for each channel and each time step. This produces a raw multi-channel observation tensor over the event region.

Because satellite observations within a 24-hour window may contain multiple valid measurements for the same spatial location, we apply a temporal compression step to organize the raw stream into a daily observation layer. For each channel and each pixel, the pipeline scans the hourly observations and retains the latest valid observation within the 24-hour window. This step removes the hourly dimension while preserving the raw sensor values, producing a tensor of shape $(C, H, W)$ for each sensor and each day. Importantly, this operation does not convert the observations into reanalysis variables or expert-derived physical products; it only organizes the raw sensor stream into a consistent event-level format.

Finally, each processed observation is saved together with its event metadata, including the disaster identifier, start date, end date, longitude, and latitude. The resulting event-centric observations are then used to construct lifecycle-oriented VQA samples covering pre-disaster anticipation, active event evolution, and post-disaster impact quantification. In total, Obshazard-bench contains 4,202 benchmark samples derived from 127 historical disaster events.

\subsection{VQA Task Generation}
\label{subsec:tasks}

After building event-centric observation samples, we convert them into visual question answering (VQA) instances. Each instance is defined by three elements: a disaster event, an observation timestep, and a target variable. This makes each sample traceable to its source event, raw observations, and ground-truth record.

For each event, we first build a temporal sequence around the event date. We use $t-14$, $t-7$, and $t-3$ for pre-event observations. These correspond to 14, 7, and 3 days before the recorded event onset. We use $t_{15}$ for event-stage observations. When available, we also use $t_{\max-1}$ and $t_{\max}$ for later event-related observations. Each timestep is paired with raw multi-channel observations from AMSU-A, HIRS, and MHS. Event metadata, such as location, event date, and disaster category, is also attached.

We then generate VQA samples by matching each timestep with eligible target variables. Pre-event timesteps are used for early-warning targets, including disaster occurrence, disaster type, arrival time, and initial duration. Event-stage and later timesteps are used for evolution and impact-related targets, including recovery time, magnitude, total deaths, total affected, number affected, number injured, number homeless, and Consumer Price Index (CPI)-related economic disruption. In this way, one disaster event can produce multiple VQA samples across different timesteps and target variables.

Each VQA instance contains four parts: visual input, question, answer space, and ground-truth answer. The visual input comes from the raw multi-channel observation tensor. The question is generated from task-specific templates and filled with event-related information. The answer space depends on the target type. Categorical targets use discrete labels. Temporal, numerical, and impact-related targets are converted into standardized answer intervals. The ground-truth answer is taken from verified disaster metadata and outcome records.

We also record metadata for each sample, including disaster category, disaster sub-category, lifecycle stage, subtask name, timestep, event identifier, and answer type. These fields are used for evaluation and analysis, not as shortcut inputs to the model. With this event--timestep--target construction, Obshazard-bench contains 4,202 VQA samples from 127 historical disaster events.

\subsection{Task Dimensions}
\label{subsec:dimensions}


\begin{table*}[t]
\centering
\small
\setlength{\tabcolsep}{4.5pt}
\renewcommand{\arraystretch}{1.18}
\caption{Lifecycle reasoning dimensions and sample distribution in ObsHazard-Bench.}
\label{tab:lifecycle_distribution}
\begin{tabularx}{0.96\textwidth}{
@{}>{\raggedright\arraybackslash}X r
   >{\raggedright\arraybackslash}X r
   >{\raggedright\arraybackslash}X r@{}}
\toprule
\multicolumn{2}{c}{\textbf{Predictive Crisis Anticipation}} &
\multicolumn{2}{c}{\textbf{Active Evolution Reasoning}} &
\multicolumn{2}{c}{\textbf{Multi-faceted Impact Quantification}} \\
\cmidrule(lr){1-2}\cmidrule(lr){3-4}\cmidrule(lr){5-6}
\textbf{Sub-task} & \textbf{Samples} &
\textbf{Sub-task} & \textbf{Samples} &
\textbf{Sub-task} & \textbf{Samples} \\
\midrule
Risk Detection & 381
& Active Termination Prediction & 213
& Magnitude Deduction & 185 \\

Type Classification & 381
& -- & --
& Humanitarian Burden & 1,670 \\

Arrival Time Prediction & 381
& -- & --
& Socio-economic Intensity & 610 \\

Initial Duration Prediction & 381
& -- & --
& -- & -- \\
\midrule
\textbf{Total} & \textbf{1,524} &
\textbf{Total} & \textbf{213} &
\textbf{Total} & \textbf{2,465} \\

\textbf{Ratio} & \textbf{36.3\%} &
\textbf{Ratio} & \textbf{5.1\%} &
\textbf{Ratio} & \textbf{58.7\%} \\
\bottomrule
\end{tabularx}
\end{table*}

Obshazard-bench defines its evaluation space through the Cartesian combination of two dimensions: the disaster taxonomy dimension and the lifecycle reasoning dimension. The disaster taxonomy dimension specifies the physical hazard type, while the lifecycle reasoning dimension specifies the operational stage and reasoning target. This design enables evaluation not only across different disaster categories, but also across different phases of disaster response.

\textbf{Disaster Taxonomy Dimension.}
The disaster taxonomy dimension follows a two-level hazard hierarchy. At the first level, Obshazard-bench covers 8 major disaster categories: Flood, Mass movement (wet), Wildfire, Earthquake, Storm, Volcanic activity, Heat-wave, and Cold-wave. At the second level, these categories are divided into 28 disaster sub-categories. Flood includes flash flood, coastal flood, general flood, and riverine flood. Mass movement (wet) includes landslide and mudslide. Wildfire includes land fire, forest fire, and general wildfire. Earthquake includes tsunami and ground movement. Storm includes hail, tornado, blizzard or winter storm, derecho, extra-tropical storm, general storm, lightning or thunderstorm, sand or dust storm, severe weather, storm surge, and tropical cyclone. Volcanic activity includes ash fall, lava flow, general volcanic activity, and pyroclastic flow. Heat-wave and Cold-wave are retained as individual sub-categories.

\textbf{Lifecycle Reasoning Dimension.}
The lifecycle reasoning dimension is designed according to the operational workflow of disaster management and is divided into three stages.

\paragraph{\textbf{Predictive Crisis Anticipation}}
This stage evaluates whether a model can identify potential disaster risks before or near the onset of an event. Representative tasks include risk detection, disaster type classification, arrival time prediction, and initial duration prediction. These tasks require the model to interpret early physical signals from raw sounding streams and associated contextual information, rather than simply recognizing visible post-event damage.

\paragraph{\textbf{Active Evolution Reasoning}}
This stage focuses on reasoning during the ongoing development of a disaster. The core task is Active Termination Prediction, which requires the model to utilize real-time sequential sounding data to forecast exactly when the ongoing event will conclude.  This setting reflects the operational need for continuous monitoring and real-time guidance, where the key question is not only what disaster is happening, but also how the event is changing over time.

\paragraph{\textbf{Multi-faceted Impact Quantification}}
This stage evaluates the model's ability to connect physical disaster signals with downstream consequences. It contains three complementary task groups.
(1) \textit{Magnitude Deduction}: inferring the intrinsic severity or intensity level of a disaster event directly from raw, multi-source observations and event-related evidence.
(2) \textit{Humanitarian Burden}: quantifying social consequences, including Total Deaths, Total Affected, Number Injured, and Number Homeless.
(3) \textit{Socio-economic Intensity}: estimating economic disruption, including impacts reflected through disaster-induced changes in consumer and regional economic indicators.
Together, these tasks test whether models can move beyond event recognition and transform raw observational evidence into consequence-oriented disaster reasoning.

By crossing the disaster taxonomy dimension with the lifecycle reasoning dimension, Obshazard-bench forms a structured evaluation matrix over hazard types and disaster-response stages. This matrix allows the benchmark to diagnose whether a model’s capability is specific to certain hazards, certain lifecycle stages, or their interaction, rather than relying only on a single overall performance score.

\subsection{Evaluation Metrics}
\label{subsec:evaluation_metrics}

Following the evaluation protocol of Earth AI~\cite{bell2026earthaiunlockinggeospatial}, Obshazard-bench adopts a task-aware scoring scheme for heterogeneous disaster reasoning tasks. Each model response is normalized into a comparable answer format before scoring. For categorical tasks, such as risk detection and disaster type classification, predictions are evaluated according to their semantic or lexical consistency with the ground-truth answer. For numerical and interval-based tasks, including arrival time prediction, duration prediction, recovery time prediction, magnitude deduction, humanitarian burden estimation, and socio-economic intensity estimation, the score is computed according to the distance between the predicted value or interval and the ground truth. All scores are normalized to the range of $[0,1]$.

To avoid bias toward sample-rich disaster categories or frequently occurring subtasks, we aggregate performance using an equal-weight protocol across timesteps, subtasks, lifecycle stages, and disaster categories. This metric design enables fair comparison across the three lifecycle stages of Obshazard-bench: Predictive Crisis Anticipation, Active Evolution Reasoning, and Multi-faceted Impact Quantification.

\subsection{Quality Control}
\label{subsec:qc}

To ensure the reliability of Obshazard-bench, we conduct quality control from both automated data construction and expert manual review.

\textbf{Automated Data Quality Control:} During data generation, we automatically check the consistency between disaster metadata, satellite observations, and constructed benchmark instances. Specifically, the pipeline verifies event time windows, location coordinates, satellite data availability, channel completeness, missing-value patterns, and the correspondence between generated samples and source records. Cases that fail these checks are removed or regenerated to ensure that each sample is grounded in valid event metadata and aligned multi-source observations.

\textbf{Expert Manual Quality Control:} In addition to automated checks, domain experts manually inspect the constructed cases and task annotations. They review whether the selected observations are relevant to the target disaster, whether the generated questions are scientifically meaningful, and whether the answers are consistent with disaster records and available evidence. Ambiguous, weakly supported, or scientifically invalid samples are revised or excluded from the final benchmark.

\section{Experiments}
\label{sec:experiments}

\begin{table*}[t]
\centering
\scriptsize
\setlength{\tabcolsep}{4pt}
\renewcommand{\arraystretch}{1.08}
\caption{\textbf{Overall performance across disaster categories.}
We report equal-weight scores for eight major disaster categories and the average score across categories. Mass Mov. denotes Mass Movement, and Avg. denotes average. The best result in each column is highlighted in bold.}
\label{tab:main_results}
\resizebox{\textwidth}{!}{
\begin{tabular}{lccccccccc}
\toprule
\textbf{Model} &
\textbf{Earthquake} &
\textbf{Flood} &
\textbf{Storm} &
\textbf{Wildfire} &
\textbf{Cold-wave} &
\textbf{Heat-wave} &
\textbf{Mass Mov.} &
\textbf{Volcanic} &
\textbf{Avg.} \\
\midrule
Claude Opus 4.8~\cite{anthropic2026claudeopus48} & 0.2252 & \textbf{0.2389} & \textbf{0.2684} & 0.2377 & 0.1574 & 0.2839 & 0.2719 & 0.2557 & 0.2424 \\
GPT-5.5~\cite{openai2026gpt55} & \textbf{0.3501} & 0.2318 & 0.2626 & \textbf{0.3681} & 0.1376 & 0.2583 & \textbf{0.3496} & \textbf{0.3081} & \textbf{0.2833} \\
Kimi-k2.6-1T~\cite{moonshot2026kimi26} & 0.2128 & 0.2344 & 0.2406 & 0.2194 & \textbf{0.2690} & \textbf{0.3022} & 0.2461 & 0.2805 & 0.2506 \\
Qwen3.5-397B-A17B~\cite{qwen2026qwen35} & 0.2091 & 0.1946 & 0.2154 & 0.1932 & 0.1422 & 0.2362 & 0.1281 & 0.2192 & 0.1922 \\
\bottomrule
\end{tabular}}
\vspace{-2mm}
\end{table*}

\subsection{Experimental Setup}
\label{subsec:experimental_setup}

We evaluate four representative foundation models on Obshazard-bench, including three closed-source frontier models and one open-source large language model. The closed-source models are Claude Opus 4.8~\cite{anthropic2026claudeopus48}, GPT-5.5~\cite{openai2026gpt55}, and Kimi-k2.6-1T~\cite{moonshot2026kimi26}; the open-source model is Qwen3.5-397B-A17B~\cite{qwen2026qwen35}. All models are evaluated under a zero-shot setting with the same input representation and uniform prompts for fair comparison. As described in Section~\ref{subsec:evaluation_metrics}, we report normalized task-aware scores following the Earth AI evaluation protocol~\cite{bell2026earthaiunlockinggeospatial}. Additional implementation details, prompt templates, and model configurations are provided in the Appendix.

\subsection{Main Results}
\label{subsec:main_results}

We evaluate representative general-purpose and domain-adapted models on Obshazard-bench, including frontier MLLMs, an open-source large language model, and a fine-tuned disaster-oriented model. All tasks are evaluated under the three lifecycle stages defined in Section~\ref{subsec:dimensions}: \textit{Predictive Crisis Anticipation} (PCA), \textit{Active Evolution Reasoning} (AER), and \textit{Multi-faceted Impact Quantification} (MIQ). Scores are normalized to $[0,1]$ and aggregated with an equal-weight protocol across timesteps, subtasks, lifecycle stages, and disaster categories, preventing the final score from being dominated by sample-rich tasks or high-frequency disaster types. Following the evaluation practice in Earth AI~\cite{bell2026earthaiunlockinggeospatial}, we use task-aware soft scoring rather than strict binary matching, where numeric answers are graded according to their distance from the ground truth and categorical or textual answers receive partial credit when semantically or lexically close.

Table~\ref{tab:main_results} reports the overall performance across the eight disaster categories. Obshazard-bench remains challenging for all evaluated models: even the best-performing model achieves an average score below 0.30. Among the evaluated general-purpose models, GPT-5.5 achieves the highest average score, followed by Kimi-k2.6-1T and Claude Opus 4.8. While GPT-5.5 provides the strongest overall performance, the remaining models exhibit complementary strengths across different disaster categories and lifecycle stages. These results show that current models still struggle to transform raw multi-channel Earth observations into lifecycle-aware disaster intelligence.

\begin{table}[t]
\centering
\scriptsize
\setlength{\tabcolsep}{5pt}
\renewcommand{\arraystretch}{1.08}
\caption{\textbf{Performance across lifecycle stages.}
PCA denotes Predictive Crisis Anticipation, AER denotes Active Evolution Reasoning, and MIQ denotes Multi-faceted Impact Quantification. The best result in each column is highlighted in bold.}
\label{tab:lifecycle_results}
\resizebox{\columnwidth}{!}{
\begin{tabular}{lcccc}
\toprule
\textbf{Model} & \textbf{PCA} & \textbf{AER} & \textbf{MIQ} & \textbf{Avg.} \\
\midrule
Claude Opus 4.8~\cite{anthropic2026claudeopus48} & \textbf{0.4697} & 0.0278 & 0.2297 & 0.2424 \\
GPT-5.5~\cite{openai2026gpt55} & 0.3443 & \textbf{0.2078} & \textbf{0.2977} & \textbf{0.2833} \\
Kimi-k2.6-1T~\cite{moonshot2026kimi26} & 0.3756 & 0.1488 & 0.2275 & 0.2506 \\
Qwen3.5-397B-A17B~\cite{qwen2026qwen35} & 0.3084 & 0.1092 & 0.1591 & 0.1922 \\
\bottomrule
\end{tabular}}
\vspace{-2mm}
\end{table}

\subsection{Capability Differences across Lifecycle Stages and Disaster Types}
\label{subsec:capability_differences}

To better understand model behavior, we analyze model performance along two benchmark dimensions: lifecycle stage and disaster type. Table~\ref{tab:lifecycle_results} shows that different models exhibit distinct lifecycle-stage capability profiles. Claude Opus 4.8 achieves the highest score on \textit{Predictive Crisis Anticipation}, indicating strong capability in pre-disaster risk detection and early forecasting. However, its performance drops sharply on \textit{Active Evolution Reasoning}, suggesting that strong pre-event reasoning does not necessarily translate into reliable reasoning about event evolution or recovery time.

GPT-5.5 shows the strongest performance on \textit{Multi-faceted Impact Quantification} and achieves the highest overall score among the evaluated models, indicating better capability in connecting event-related evidence with consequence-level estimates. Kimi-k2.6-1T exhibits a capability profile between GPT-5.5 and Claude Opus 4.8: it performs strongly on \textit{Predictive Crisis Anticipation} and moderately on \textit{Multi-faceted Impact Quantification}, but remains limited on \textit{Active Evolution Reasoning}. Claude Opus 4.8 performs competitively on \textit{Predictive Crisis Anticipation}, while Qwen3.5-397B-A17B remains consistently behind the stronger closed-source models. These results indicate that model capacity and general reasoning capability remain important for raw-observation-based disaster intelligence.

These lifecycle-stage results suggest that disaster intelligence is not a single homogeneous capability. PCA emphasizes early physical signal interpretation, AER requires temporal evolution modeling, and MIQ requires connecting physical observations with humanitarian and socio-economic consequences. A model that performs well in one stage may still fail in another, demonstrating the necessity of evaluating the full disaster lifecycle rather than reporting only a single aggregated score.

The disaster-level results in Table~\ref{tab:main_results} further reveal strong model--hazard interactions. GPT-5.5 performs best on earthquake, wildfire, and mass movement events, which may benefit from broader world knowledge and semantic reasoning about disaster mechanisms. Claude Opus 4.8 achieves the best score on storm, while Qwen3.5-397B-A17B shows weaker performance across most disaster categories. These results demonstrate that model capability varies substantially across hazard families, even under the same benchmark format.

These results demonstrate that model capability varies substantially across hazard families. A model that performs well on one disaster type may not generalize to another, even under the same benchmark format. This supports the need for a multi-hazard benchmark such as Obshazard-bench, where evaluation spans hydrological, meteorological, geophysical, and environmental disasters rather than focusing on a single hazard family. Overall, the lifecycle-stage and disaster-type analyses jointly show that current MLLMs exhibit structured capability gaps across both operational stages and physical hazard mechanisms.

\begin{figure*}[t]
    \centering
    \includegraphics[width=0.95\textwidth]{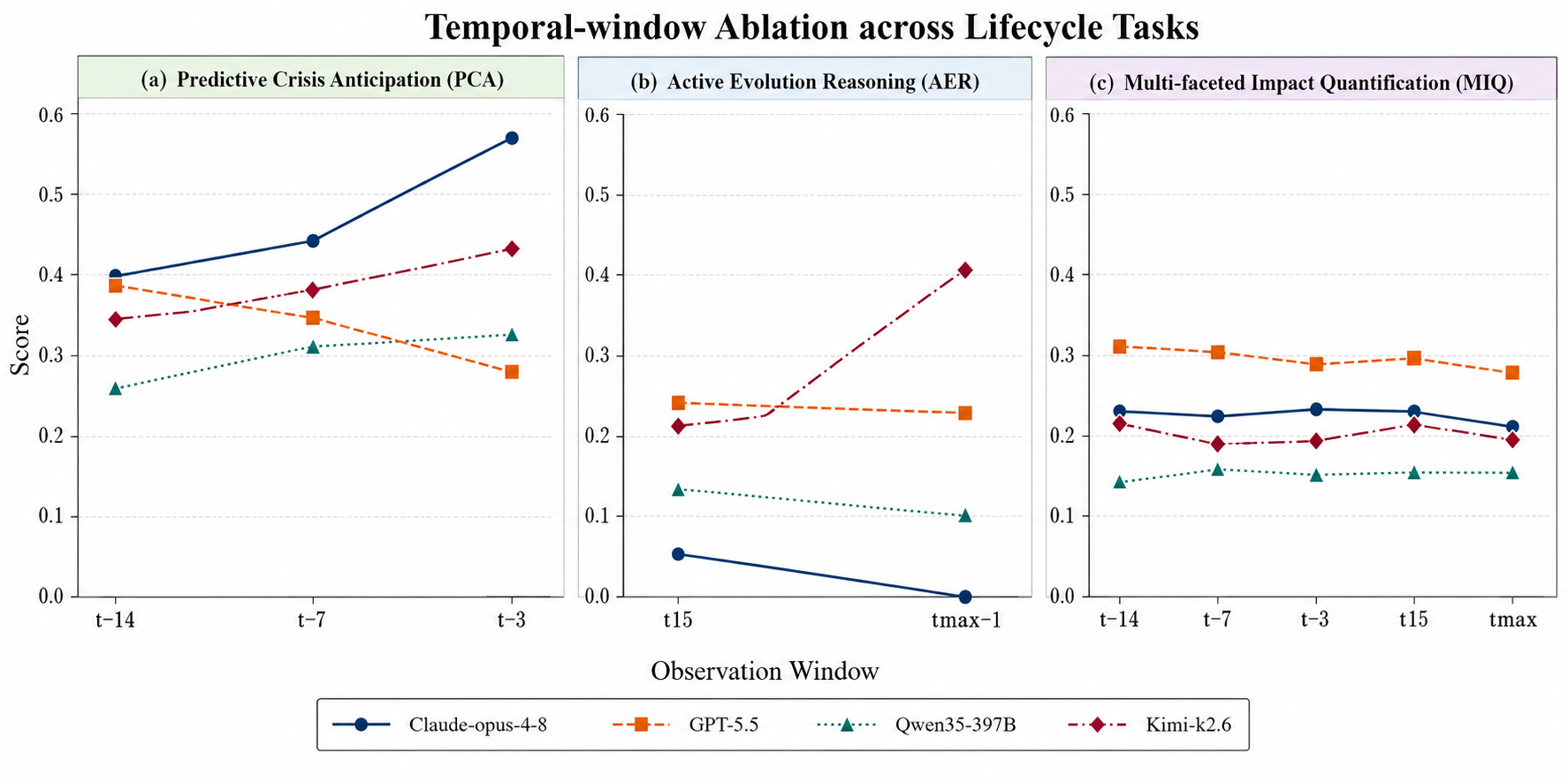}
    \caption{\textbf{Temporal-window ablation across lifecycle tasks.}
    We compare model performance under different observation windows for three lifecycle tasks: Predictive Crisis Anticipation (PCA), Active Evolution Reasoning (AER), and Multi-faceted Impact Quantification (MIQ). PCA, AER, and MIQ denote Predictive Crisis Anticipation, Active Evolution Reasoning, and Multi-faceted Impact Quantification, respectively.}
    \label{fig:temporal_window_ablation}
\end{figure*}

\subsection{Temporal-window Ablation Study}
\label{subsec:temporal_ablation}

\begin{table}[t]
\centering
\scriptsize
\setlength{\tabcolsep}{7pt}
\renewcommand{\arraystretch}{1.08}
\caption{\textbf{Temporal-window ablation for Predictive Crisis Anticipation.}
We report PCA scores using observations collected 14, 7, and 3 days before event onset.}
\label{tab:temporal_pca}
\resizebox{\columnwidth}{!}{
\begin{tabular}{lccc}
\toprule
\textbf{Model} & \multicolumn{3}{c}{\textbf{Days before event onset}} \\
\cmidrule(lr){2-4}
& \textbf{14} & \textbf{7} & \textbf{3} \\
\midrule
Claude Opus 4.8~\cite{anthropic2026claudeopus48} & 0.3973 & \textbf{0.4385} & \textbf{0.5731} \\
GPT-5.5~\cite{openai2026gpt55} & \textbf{0.3893} & 0.3529 & 0.2908 \\
Kimi-k2.6-1T~\cite{moonshot2026kimi26} & 0.3750 & 0.3453 & 0.4065 \\
Qwen3.5-397B-A17B~\cite{qwen2026qwen35} & 0.2717 & 0.3204 & 0.3331 \\
\bottomrule
\end{tabular}}
\vspace{-2mm}
\end{table}

\begin{table}[t]
\centering
\scriptsize
\setlength{\tabcolsep}{8pt}
\renewcommand{\arraystretch}{1.08}
\caption{\textbf{Temporal-window ablation for Active Evolution Reasoning.}
We report AER scores using two event-stage observation checkpoints: the event-day checkpoint and a late-stage checkpoint before the maximum timestep.}
\label{tab:temporal_aer}
\resizebox{\columnwidth}{!}{
\begin{tabular}{lcc}
\toprule
\textbf{Model} & \multicolumn{2}{c}{\textbf{Event-stage observation checkpoint}} \\
\cmidrule(lr){2-3}
& \textbf{Start day} & \textbf{End-1 day} \\
\midrule
Claude Opus 4.8~\cite{anthropic2026claudeopus48} & 0.0557 & 0.0000 \\
GPT-5.5~\cite{openai2026gpt55} & \textbf{0.2443} & \textbf{0.2318} \\
Kimi-k2.6-1T~\cite{moonshot2026kimi26} & 0.1322 & 0.1842 \\
Qwen3.5-397B-A17B~\cite{qwen2026qwen35} & 0.1270 & 0.0925 \\
\bottomrule
\end{tabular}}
\vspace{-2mm}
\end{table}

\begin{table}[t]
\centering
\scriptsize
\setlength{\tabcolsep}{5pt}
\renewcommand{\arraystretch}{1.08}
\caption{\textbf{Temporal-window ablation for Multi-faceted Impact Quantification.}
We report MIQ scores across five observation checkpoints, spanning pre-event lead times and event-stage observations.}
\label{tab:temporal_miq}
\resizebox{\columnwidth}{!}{
\begin{tabular}{lccccc}
\toprule
\textbf{Model} & \multicolumn{3}{c}{\textbf{Days before event onset}} & \multicolumn{2}{c}{\textbf{Event-stage observation checkpoint}} \\
\cmidrule(lr){2-4}\cmidrule(lr){5-6}
& \textbf{14} & \textbf{7} & \textbf{3} & \textbf{Start day} & \textbf{End day} \\
\midrule
Claude Opus 4.8~\cite{anthropic2026claudeopus48} & 0.2366 & 0.2280 & 0.2379 & 0.2329 & 0.2131 \\
GPT-5.5~\cite{openai2026gpt55} & \textbf{0.3138} & \textbf{0.3058} & \textbf{0.2916} & \textbf{0.2978} & \textbf{0.2794} \\
Kimi-k2.6-1T~\cite{moonshot2026kimi26} & 0.2440 & 0.2127 & 0.2357 & 0.2324 & 0.2128 \\
Qwen3.5-397B-A17B~\cite{qwen2026qwen35}& 0.1475 & 0.1664 & 0.1592 & 0.1609 & 0.1614 \\
\bottomrule
\end{tabular}}
\vspace{-2mm}
\end{table}

We further conduct a temporal-window ablation study to examine how model performance changes when observations are taken at different timesteps. Unlike the previous analysis, which focuses on model capability differences across stages and hazard types, this study investigates whether different lifecycle tasks benefit from observations closer to the disaster event. This is important for raw-stream disaster intelligence, where practical systems must decide how much historical observation to retain, how frequently to run inference, and which time windows are most informative for different operational tasks.

Table~\ref{tab:temporal_pca} shows the ablation results for \textit{Predictive Crisis Anticipation} using observations collected 14, 7, and 3 days before event onset. Claude Opus 4.8 increases from 0.3973 with a 14-day lead time to 0.5731 with a 3-day lead time, suggesting that near-onset observations can contain more informative physical signals for early-warning tasks. However, GPT-5.5 shows a counterintuitive decline from 0.3893 to 0.2908 as the lead time shortens. Kimi-k2.6-1T follows a non-monotonic pattern, dropping at the 7-day lead time and recovering at the 3-day lead time, suggesting that its temporal sensitivity is weaker and less stable than Claude Opus 4.8. This indicates that temporal proximity alone does not guarantee improved reasoning if a model cannot effectively integrate sequentially updated observations. The result highlights the value of evaluating multiple pre-event lead times rather than relying on a single static pre-disaster snapshot.

Table~\ref{tab:temporal_aer} further evaluates \textit{Active Evolution Reasoning} using event-stage observations. GPT-5.5 performs best at both checkpoints and remains stable at the late-stage checkpoint, while Claude Opus 4.8 obtains near-zero performance at the late-stage checkpoint. Kimi-k2.6-1T improves from the event-day checkpoint to the late-stage checkpoint, indicating a mild benefit from additional event-tail observations. This highlights the importance of temporal grounding for reasoning about disaster evolution and recovery.

Table~\ref{tab:temporal_miq} shows a different pattern for \textit{Multi-faceted Impact Quantification}. Across all models, MIQ performance remains relatively stable across timesteps. Kimi-k2.6-1T follows the same broadly time-invariant pattern, with only small fluctuations across the five checkpoints. This suggests that consequence-level estimation is less sensitive to temporal proximity than early-warning tasks. One possible explanation is that humanitarian burden and socio-economic intensity cannot be inferred from physical observations alone; they also require robust grounding in event metadata, exposure, vulnerability, and historical context. Therefore, additional observations closer to the event do not automatically improve consequence-level estimation.

Overall, the temporal-window ablation reveals that different lifecycle tasks respond differently to observation timing. PCA is generally time-sensitive and often benefits from near-onset signals, MIQ is relatively time-invariant and likely requires additional contextual grounding beyond raw observations, while AER can benefit from late-stage event-tail evidence. These findings demonstrate that Obshazard-bench can evaluate not only whether a model performs well, but also when observations become useful for different types of disaster reasoning.

\subsection{Operational Implications}
\label{subsec:operational_implications}

Beyond model ranking, the experimental results provide several implications for practical disaster-intelligence workflows. First, the temporal trend in PCA suggests the possibility of adaptive observation-window retention. If future models can reliably improve as event onset approaches, operational systems may preserve dense observations only within shorter high-risk rolling windows, reducing storage and inference costs while maintaining early-warning utility. However, this should be balanced against the value of longer lead time, since earlier but less accurate predictions may still be useful for evacuation planning and emergency resource pre-positioning.

Second, the weak temporal sensitivity of MIQ indicates that consequence-level estimation may not require continuous inference at every observation timestep. In practice, MIQ modules could be executed at selected checkpoints, while more system resources are allocated to collecting exposure, vulnerability, infrastructure, and socio-economic context. This suggests that improving impact assessment may depend less on simply increasing raw observation frequency and more on integrating decision-relevant contextual information.

Third, the results show that no single evaluated model performs consistently well across all lifecycle stages and disaster categories. General-purpose models, although strong in some settings, still exhibit clear stage-specific and hazard-specific capability gaps. This suggests that raw-observation-based disaster intelligence may involve multiple technical bottlenecks rather than a single unified reasoning ability. From an operational perspective, future disaster-response systems may benefit from combining complementary models or modules, where different models specialize in early warning, event evolution, or impact quantification.

Finally, given the low absolute scores, current MLLMs should be viewed as decision-support tools rather than autonomous disaster-response agents. For high-stakes tasks such as evacuation planning, humanitarian burden estimation, and recovery scheduling, model outputs should be coupled with uncertainty thresholds, expert review, and escalation mechanisms.

Overall, the experimental results show that current MLLMs still struggle to transform raw multi-channel Earth observations into lifecycle-aware disaster intelligence. The failures are not uniform: they depend on the lifecycle stage, temporal distance to the event, and disaster type. Obshazard-bench thus serves not only as a leaderboard, but also as a diagnostic benchmark for identifying stage-specific, time-dependent, and hazard-dependent limitations in current foundation models.

\section{Conclusion}
\label{sec:conclusion}

We present Obshazard-bench, a real-time observation-driven benchmark for evaluating disaster intelligence from raw Earth observation streams. By integrating satellite sounding observations, disaster metadata, and auxiliary contextual information, Obshazard-bench covers 8 major disaster categories and 28 sub-categories across 127 historical disaster events. Furthermore, we introduce a lifecycle-oriented evaluation taxonomy consisting of \textit{Predictive Crisis Anticipation}, \textit{Active Evolution Reasoning}, and \textit{Multi-faceted Impact Quantification}, enabling systematic assessment of disaster reasoning before, during, and after event occurrence.

Experimental results show that current MLLMs remain far from solving raw-observation-based disaster intelligence. Performance varies substantially across lifecycle stages, disaster categories, and observation windows, revealing distinct temporal, hazard-specific, and reasoning-related capability gaps. These findings suggest that disaster intelligence cannot be treated as a single homogeneous capability and requires stronger temporal grounding, event-evolution modeling, and consequence-level reasoning.

We hope Obshazard-bench can serve as a standardized testbed for future disaster foundation models and stimulate research on real-time Earth observation understanding, lifecycle-aware disaster reasoning, and decision-oriented geospatial intelligence.

\bibliographystyle{IEEEtran}
\bibliography{ref}

\end{document}